\documentclass{article}

\usepackage[final]{nips_2016}

\usepackage[utf8]{inputenc} 
\usepackage[T1]{fontenc}    
\usepackage{hyperref}       
\usepackage{url}            
\usepackage{booktabs}       
\usepackage{amsfonts}       
\usepackage{nicefrac}       
\usepackage{microtype}      

\usepackage{mathtools}
\usepackage{threeparttable}
\usepackage{graphicx}
\usepackage{xcolor}

\title{CORe50: a New Dataset and Benchmark for Continuous Object Recognition}

\author{
  Vincenzo Lomonaco \& Davide Maltoni\\
  Department of Computer Science and Engineering - DISI\\
  Alma Mater Studiorum - University of Bologna\\
  \url{http://vlomonaco.github.io/core50} \\
  \texttt{\{vincenzo.lomonaco, davide.maltoni\}@unibo.it} \\
}

\begin{document}

\maketitle

\begin{abstract}
Continuous/Lifelong learning of high-dimensional data streams is a challenging research problem. In fact, fully retraining models each time new data become available is infeasible, due to computational and storage issues, while naïve incremental strategies have been shown to suffer from catastrophic forgetting. In the context of real-world object recognition applications (e.g., robotic vision), where continuous learning is crucial, very few datasets and benchmarks are available to evaluate and compare emerging techniques. In this work we propose a new dataset and benchmark \textsl{CORe50}, specifically designed for continuous object recognition, and introduce baseline approaches for different continuous learning scenarios.
\end{abstract}

\section{Introduction}

Datasets such as \textsl{ImageNet} and \textsl{Pascal VOC} provide a very good playground for classification and detection approaches. However, they have been designed with “static” evaluation protocols in mind; the entire dataset is split in just two parts: a training set is used for (one-shot) learning and a separate test set is used for accuracy evaluation. 
Splitting the training set into a number of batches is essential to train and test continuous learning approaches, a hot research topic that is currently receiving much attention. Unfortunately, most of the existing datasets are not well suited to this purpose because they lack a fundamental ingredient: the presence of multiple (unconstrained) views of the same objects taken in different sessions (varying background, lighting, pose, occlusions, etc.).
Focusing on Object Recognition we consider three continuous learning scenarios:
\begin{itemize}
\item \textbf{New Instances} (\textbf{NI}): new training patterns of the same classes become available in subsequent batches with new poses and conditions (illumination, background, occlusion, etc.). A good model is expected to incrementally consolidate its knowledge about the known classes without compromising what it has learned before.
\item \textbf{New Classes} (\textbf{NC}): new training patterns belonging to different classes become available in subsequent batches. In this case the model should be able to deal with the new classes without losing accuracy on the previous ones. 
\item \textbf{New Instances and Classes} (\textbf{NIC}): new training patterns belonging both to known and new classes become available in subsequent training batches. A good model is expected to consolidate its knowledge about the known classes and to learn the new ones.
\end{itemize}

It appears clear that for dealing with such complex scenarios datasets/benchmarks specifically designed for continuous learning are needed in order to evaluate and compare emerging approaches. In particular, the ideal dataset should be composed of a high number of classes to be added in subsequent batches, but, more important, of a large number of views for each class acquired in different sessions. The presence of temporal coherent sessions (i.e., videos where the objects gently move in front of the camera) is another key feature since temporal smoothness can be used to simplify object detection, improve classification accuracy and to address unsupervised scenarios \	\cite{Maltoni2016-icpr} \cite{Maltoni2016-arxiv}. In Section \ref{sec:related} we review existing datasets and better clarify the motivations that induced us to propose a new benchmark.
When addressing real-world continuous learning, assuming that the whole past data can be used at each training step (i.e., cumulative approach) is not only very far from biological learning but also unlikely for application engineering. In fact, the cumulative approach would require:
\begin{itemize}
\item To store all the previous data streams;
\item To retrain a model on all the whole data each time new data is available. 
\end{itemize}

Updating an already trained model with new data (only) is much more feasible in term of computation and memory constraints. Recent advances in transfer learning/tuning with deep neural networks have shown that using previously learned knowledge on similar tasks can be useful for solving new ones \cite{Lomonaco2016}. Yet, little has been done in the context of continuous learning where the same model is required to solve new tasks while maintaining good performances on the previous ones. Indeed, preserving previously learned knowledge without re-accessing old patterns remains particularly challenging due to the phenomenon known in literature as catastrophic forgetting where what has been learned so far is seriously compromised.
The contributions of this paper can be summarized as follows:
\begin{itemize}
\item Collection of a new dataset (called \textsl{CORe50}) specifically designed for continuous object recognition. \textsl{CORe50} is publicly available at \url{http://vlomonaco.github.io/core50}.
\item Definition of benchmarks for NI, NC, NIC scenarios on \textsl{CORe50}.
\item Design and test of simple continual learning approaches for NI, NC, NIC scenarios to be used as baseline references when developing new approaches.
\end{itemize}

In section \ref{sec:related} we review related literature and compare datasets that can be used for continuous object recognition. Section \ref{sec:dataset} describes \textsl{CORe50}. While the new dataset is intended for continuous object recognition, in Section \ref{sec:static-bench} we provide an overview of the accuracy that can be achieved by training recent CNNs on the whole training data (i.e., static benchmark). In Section \ref{sec:cont-bench} we introduce, for each continuous learning scenario (NI, NC and NIC), one or more simple approaches that can be used as baseline references when testing new developments on \textsl{CORe50}. Finally, in Section \ref{sec:conclusion} some conclusions are drawn.

\section{Related Works}
\label{sec:related}

In the last few years, we have witnessed a renewed interest in continuous learning, both for supervised classification and reinforcement learning. Several interesting approaches have been proposed such as: Learning without Forgetting \cite{Li2016}, Progressive Neural Networks \cite{Rusu}, Active Long Term Memory Networks \cite{Furlanello2016}, Adaptive Convolutional Neural Network \cite{Zhang2016}, PathNet \cite{pathnet}, Incremental Regularized Least Squares \cite{Camoriano2017}, Elastic Weight Consolidation \cite{Kirkpatrick2016}, Encoder-based Lifelong Learning \cite{Triki2017}, etc.

\textsl{Elastic Weight Consolidation} (\textsl{EWC}) appears particularly intriguing since it provides a formal criterion to identify the subset of weights that should not be changed (or changed as little as possible) during the model update to reduce forgetting. \textsl{Learning without Forgetting} (\textsl{LwF}) \cite{Li2016} and some of its evolutions \cite{Triki2017} are also very interesting since they prove that enforcing the output stability helps to control forgetting and, at the same time, provides enough degrees of freedom to learn the new task(s).     
Unfortunately, until now, most of the experimentations have been performed in setups that are appropriate to learn very short sequences of big tasks (typically 2), while their capabilities have not been assessed in continuous learning scenarios characterized by frequent (sometimes small) updates. For example, \textsl{EWC} classification tests have been carried out on permuted \textsl{MNIST} task \cite{Goodfellow2013} \cite{Srivastava2013}, where the new tasks to learn are obtained by scrambling the pixel positions in the \textsl{MNIST} dataset. Although original and ingenious such experiment is not truly linked to real-world applications. Most of the experiments performed on \textsl{LwF} consider pairs of large datasets (selected from \textsl{ImageNet}, \textsl{Pascal VOC}, \textsl{Places365-standard}, \textsl{Caltech-UCSD} \textsl{Birds-200-2011}, \textsl{MIT indoor scene classification}, etc) and proves that a model trained on the first dataset can be further trained on the second one without forgetting the first task; even if \textsl{LwF} can deal with multiple new tasks, only a simple experiments is reported in \cite{Li2016} for multiple new tasks. Furthermore, only the NC scenario has been addressed.

In Table \ref{tab:datasets} we compare datasets/benchmarks which, in our opinion, could be used for continuous object recognition. Datasets where temporal coherent sequences are not available (or cannot be generated from static frames) are here excluded. In principle, such datasets could be used for continuous learning as well, but we think that temporally coherent sequences allow a larger number of real-world applications to be addressed (e.g., robotic vision scenario). \textsl{YouTube-8M} \cite{Abu-El-Haija2016} provides a huge number of videos acquired in difficult natural settings. However, the classes are quite heterogeneous and acquisition conditions are completely uncontrolled in terms of object distance, pose, lighting, occlusions, etc. In other words, we believe it is too challenging for current continuous learning approaches (still in their infancy). 

\begin{table}[t]
  \caption{Comparison of datasets (with temporal coherent sessions) for continuous object recognition.}
  \label{tab:datasets}
  \centering
  \begin{threeparttable}
  \begin{tabular}{p{3.2cm}p{0.5cm}p{0.6cm}p{0.5cm}p{1.3cm}p{1.2cm}p{1.8cm}p{1.3cm}}
    \toprule
    Dataset & Cat.  & Obj. & Sess. & Frames per sess. & Format & Acquisition setting & Outdoor sessions\\
    \midrule
    NORB \cite{LeCun2004} & 5 & 25 & 20 & 20\tnote{*} & grayscale & turntable & no\\
    COIL-100 \cite{Nene1996} & - & 100 & 20 & 54\tnote{*} & RGB & turntable & no\\
	iLab-20M \cite{Borji2016} & 15 & 704 & - & - & RGB & turntable & no\\
	RGB-D \cite{Schwarz2015} & 51 & 300 & - & - & RGB-D & turntable & no\\
	BigBIRD \cite{bigbird} & - & 100 & - & - & RGB-D & turntable & no\\
	ALOI \cite{aloi} & - & 1000 & - & - & RGB & turntable & no\\
	BigBrother \cite{Franco2009} & - & 7 & 54 & $\sim$20 & RGB & wall cameras & no\\
	iCubWorld28 \cite{Pasquale2015a} & 7  & 28 & 4 & $\sim$150 & RGB & hand hold & no\\
	iCubWorld-Transf \cite{icub-transf} & 15 & 150 & 6 & $\sim$150 & RGB & hand hold & no\\
	\midrule
	\textbf{CORe50} & \textbf{10} & \textbf{50} & \textbf{11} & \textbf{$\sim$300} & \textbf{RGB-D} & \textbf{hand hold} & \textbf{yes (3)}\\
    \bottomrule
  \end{tabular}
  \begin{tablenotes}
    \vspace{5pt}
    \begin{footnotesize}
	\item[*] Temporal coherent training/test sessions for NORB and COIL-100 have been defined in \cite{Maltoni2016-icpr} and \cite{Maltoni2016-arxiv}.
	\end{footnotesize}
	\end{tablenotes}
  \end{threeparttable}
\end{table}

In the first group of datasets in Table \ref{tab:datasets} (\textsl{NORB}, \textsl{COIL-100}, \textsl{iLAB20M}, \textsl{Washington RGB-D}, \textsl{BigBIRD}, \textsl{ALOI}), objects are positioned on turntables and acquisition is systematically controlled in term of pose/lighting. Neither complex backgrounds nor occlusions are present in these datasets. For \textsl{NORB} and \textsl{COIL-100} we defined in \cite{Maltoni2016-icpr} a number of exploration sequences that turn the native static benchmarks into continuous learning tasks; \cite{Maltoni2016-icpr} also reports supervised and semi-supervised accuracy for the NI scenario. Exploration sequences can be generated for the other datasets in this group as well by randomly walking through adjacent static frames in the multivariate parameter space; however, the obtained sequences would remain quite unnatural.   
\textsl{BigBrother} dataset (see \cite{Lomonaco2016} \cite{Franco2009}) is an interesting incremental learning setup in the face recognition domain, but unfortunately the owner copyright does not allow the public diffusion of the dataset. 
Finally, the \textsl{iCubWorld} datasets (\cite{Pasquale2015a}, \cite{icub-transf}) have been acquired in a robotic vision context and are the closest ones to \textsl{CORe50}. In fact, objects are hand hold at nearly constant distance from the camera and are randomly moved. With respect to these datasets, \textsl{CORe50} consists of a higher number of longer sessions (including outdoor ones), more complex backgrounds and also provide depth information (that can be used as extra-feature for classification and/or to simplify object detection).

 In our opinion, the most important feature of \textsl{CORe50}, is the presence of 11 distinct acquisition sessions per object; this allows to define incremental strategies that are long enough to appreciate the learning trends.
While preparing this paper we noted that \textsl{iCubWorld-Transf} is being expanded (see \url{https://robotology.github.io/iCubWorld/} for latest updates), and we think that cross-evaluating continuous object recognition approaches on both \textsl{CORe50} and \textsl{iCubWorld-Transf} could be very interesting.

\begin{figure}[ht]
  \centering
  \includegraphics[width=0.8\textwidth]{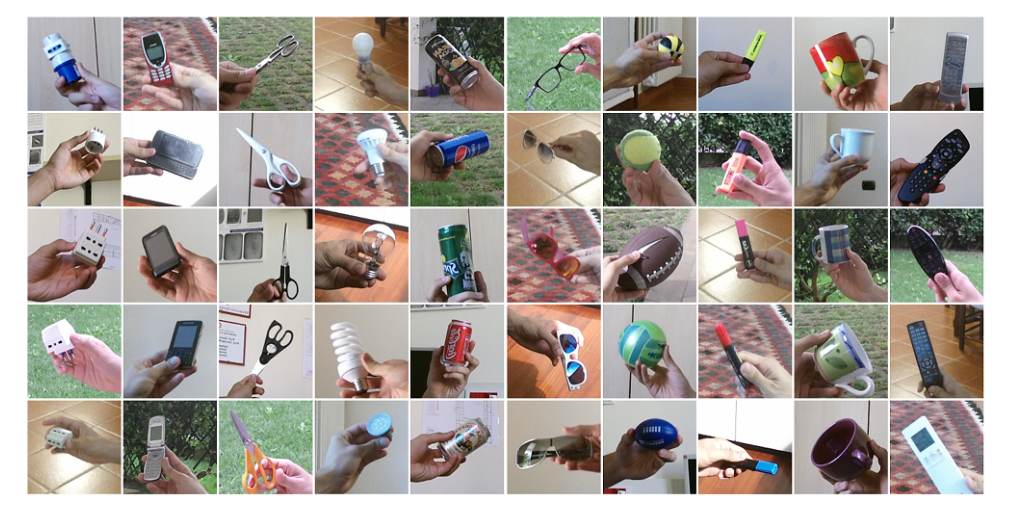}
  \caption{Example images of the 50 objects in \textsl{CORe50}. Each column denotes one of the 10 categories.}
  \label{img:core50}
\end{figure}

\section{CORe50}
\label{sec:dataset}

\textsl{CORe50}, specifically designed for (C)ontinuous (O)bject (Re)cognition, is a collection of 50 domestic objects belonging to 10 categories: plug adapters, mobile phones, scissors, light bulbs, cans, glasses, balls, markers, cups and remote controls (see Figure \ref{img:core50}). Classification can be performed at object level (50 classes) or at category level (10 classes). The first task (the default one) is much more challenging because objects of the same category are very difficult to be distinguished under certain poses.  
The dataset has been collected in 11 distinct sessions (8 indoor and 3 outdoor) characterized by different backgrounds and lighting. For each session and for each object, a 15 seconds video (at 20 fps) has been recorded with a Kinect 2.0 sensor \cite{kinect} delivering 300 RGB-D frames. 
Objects are hand hold by the operator and the camera point-of-view is that of the operator eyes. The operator is required to extend his arm and smoothly move/rotate the object in front of the camera. A subjective point-of-view with objects at grab-distance is well-suited for a number of robotic applications. The grabbing hand (left or right) changes throughout the sessions and relevant object occlusions are often produced by the hand itself.

\begin{figure}[h]
  \centering
  \includegraphics[width=0.70\textwidth]{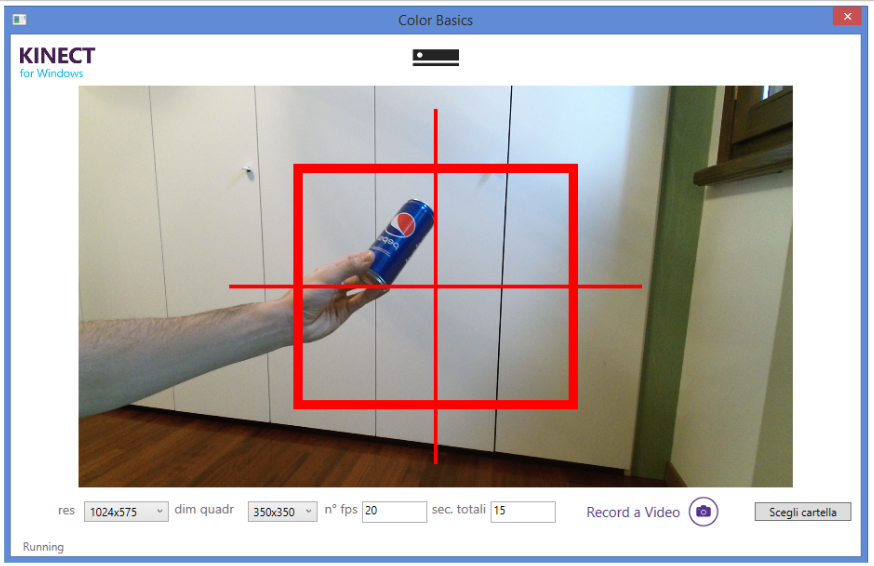}
  \caption{Acquisition interface: the red box identifies the central region where the operator is required to keep the objects while moving and rotating them.}
  \label{img:acquisition}
\end{figure}

Row data consists of 1024 $\times$ 575 RGB + 512 $\times$ 424 Depth frames. Depth information can be mapped to RGB coordinates upon calibration. The acquisition interface identifies a central region where the object should be kept (see red box in Figure \ref{img:acquisition}). This allows to performs a first (fixed) cropping, thus reducing the frame size to 350 $\times$ 350. 

\begin{figure}[h]
  \centering
  \includegraphics[width=\textwidth]{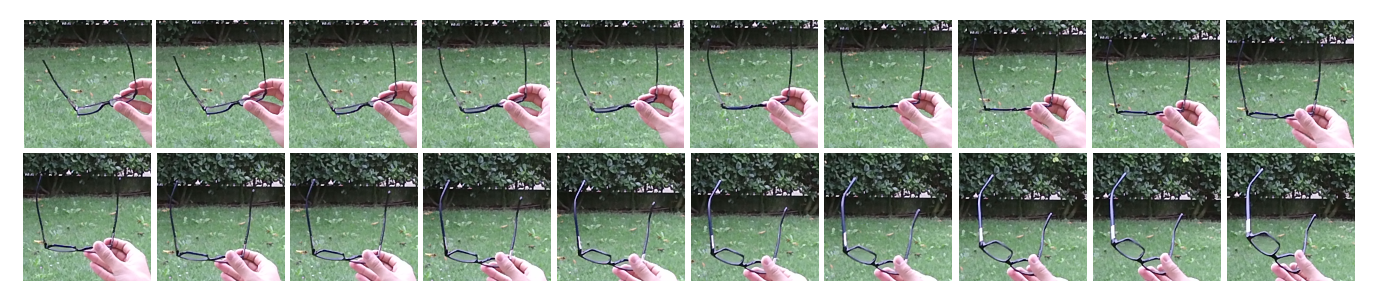}
  \caption{Example of 1 second recording (at 20 fps) of object \#26 in session \#4 (outdoor). Note the smooth movement, pose change and partial occlusion. The 128 $\times$ 128 frames here shown have been automatically cropped from 350 $\times$ 350 images based on a fully automated tracker.}
  \label{img:seq}
\end{figure}

Since our domestic objects (kept at arm distance) typically extend for less than 100 $\times$ 100 pixels, only a small fraction of the frame contains the object of interest. Therefore, we exploited temporal information to crop from each 350 $\times$ 350 frame a 128 $\times$ 128 box around the object\footnote{in principle object detection inside the frame could be performed by a trained model like a Faster R-CNN \cite{Ren2015} or Yolo \cite{Redmon2016}; however, since our main interest is continuous object recognition, at this stage we preferred to focus on object classification instead of the more complex and time demanding detection task.}. To this purpose we implemented a simple but effective motion-based tracker working only on RGB data, so that a similar approach could be used even if depth information is not available (see Figure \ref{img:seq} for an example). While in most of the cases the objects are fully contained in the crop window, sometimes they can extend beyond borders (e.g., this can happen if the object distance from the camera is reduced too much, or the tracker partially loses the object because of a too fast movement). No manual correction has been applied, because we believe that tracking imperfections are unavoidable and should be properly dealt with at later processing stages.  

The final dataset consists of 164,866 128 $\times$ 128 RGB-D images: 11 sessions x 50 objects x ($\sim$300\footnote{Some sequences are slightly shorter than 300 frames because a few initial frames are necessary to initialize the automatic motion-based tracker.}) frames per session. 
Figure \ref{img:seq} shows one frame of the same object throughout the eleven sessions. Three of the eleven sessions (\#3, \#7 and \#10) have been selected for test and the remaining 8 sessions are used for training. We tried to balance as much as possible the difficulty of training and test sessions with respect to: indoor/outdoor, holding hand (left or right) and complexity of the background. 

\begin{figure}[h]
  \centering
  \includegraphics[width=0.8\textwidth]{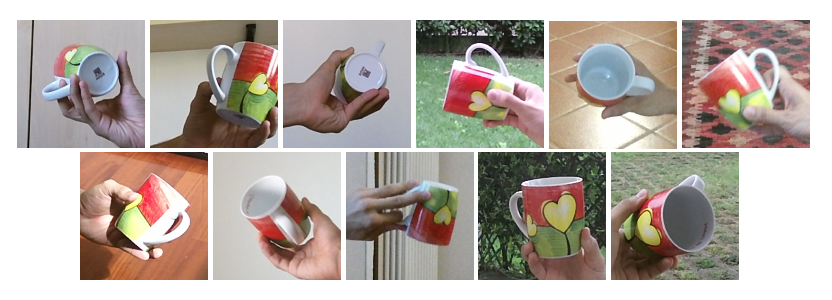}
  \caption{One frame of the same object (\#41) throughout the 11 acquisition sessions. Note the variability in terms of background, illumination, blurring, occlusion, pose and scale.}
\end{figure}

The full dataset, along with further information can be downloaded from \url{vlomonaco.github.io/core50}. In the same repository we make available the code for the reproducibility of the benchmarks described in the following sections.

\section{Static Object Recognition Benchmark}
\label{sec:static-bench}

While designed for continuous learning, \textsl{CORe50} dataset can still be used as a medium size benchmark for object recognition with a static evaluation protocol. The high object pose variability and complex acquisition setting make the problem sufficiently hard to solve even when learning is performed on the whole training data. 

In Table \ref{tab:static-core50}, we show the accuracy of two well-known CNN models (CaffeNet and VGG\footnote{We refer to the VGG-CNN-M model introduced in \cite{Chatfield2014}.}) adapted to medium size and trained in three different modalities by using RGB data only (depth information will be used in future studies):
\begin{enumerate}
\item Mid-CNN from scratch: training a model from scratch.
\item Mid-CNN + SVM: using a model pre-trained on \textsl{ILSVRC-2012} as a fixed feature extractor in conjunction with a linear SVM classifiers. Features are extracted at \textit{pool5} level.
\item Mid-CNN + FT: fine-tuning an \textsl{ILSVRC-2012} pre-trained model on \textsl{CORe50}.  
\end{enumerate}

As already shown by many authors, fine-tuning a pre-trained model on the new dataset is often the most effective strategy, especially if the new dataset is large enough to avoid overfitting, but not so large to learn representative features from scratch.

\begin{table}[ht]
  \caption{Accuracy of CaffeNet and VGG models (both adapted to size 128 $\times$ 128) on \textsl{CORe50} for different learning strategies. The test set consists of sessions: \#3, \#7 and \#10; the training set of the remaining 8 sessions.}
  \label{tab:static-core50}
  \centering
  \begin{tabular}{lcccc}
    \toprule
   	 & \multicolumn{2}{c}{\textbf{Accuracy \%}} & \multicolumn{2}{c}{\textbf{Accuracy \%}}\\
   	 & \multicolumn{2}{c}{\textbf{(object level: 50 classes)}} & \multicolumn{2}{c}{\textbf{(category level: 10 classes)}}\\
    \cmidrule{2-5}
     \textbf{Strategy} &  \textbf{CaffeNet} & \textbf{VGG} & \textbf{CaffeNet} & \textbf{VGG}\\
    \toprule
    Mid-CNN from scratch & 37,82\% & 38,09\% & 48,93\% & 53,74\%\\
    Mid-CNN + SVM & 51,35\%  & 59,03\% & 61,81\% & 68,94\%\\
    Mid-CNN + FT & 65,98\% & 69,08\% & 77,76\% & 80,23\%\\
    \bottomrule
  \end{tabular}
\end{table}

The term Mid-CNN is here used to highlight that we are not using the original 227 $\times$ 227 CaffeNet and 224 $\times$ 224 VGG models but their adaption to a mid-size of 128 $\times$ 128 pixels. Many researchers use available pre-trained CNN models as they are, and simply stretch their images to fit the model input size, even if the image size is much smaller than the CNN input. Stretching our input pattern (from 128 $\times$ 128 to 227 $\times$ 227) would require much more computation at inference time (about four times), so we decided to adapt the pre-trained CNN models to work with 128 $\times$ 128 input images. However, in case of pre-trained models, this step is not neutral and obvious as one could expect: more details are provided in Appendix \ref{appndx:mid-size}.

\begin{figure}[h]
  \centering
  \includegraphics[width=0.55\textwidth]{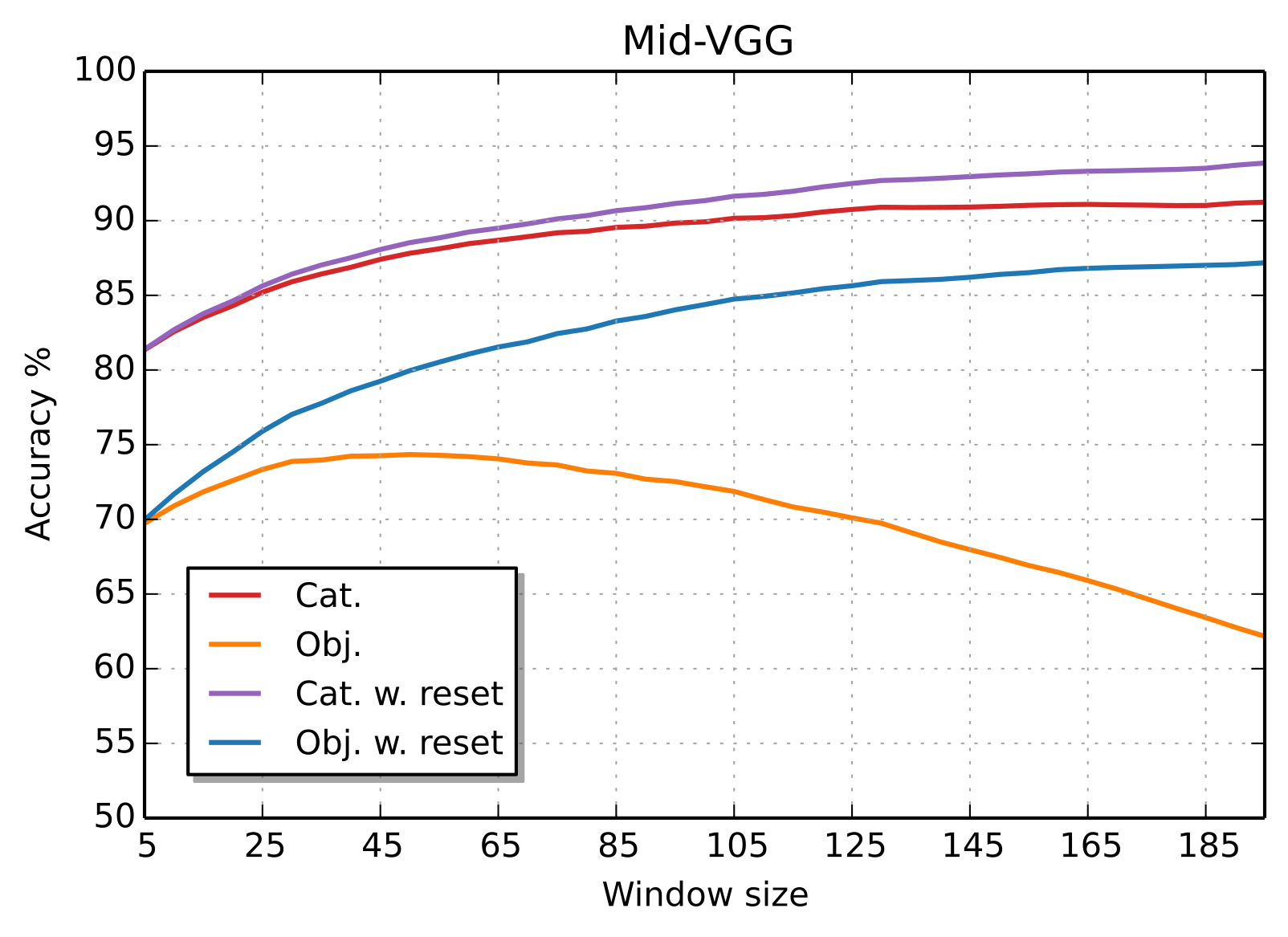}
  \caption{Mid-VGG classification accuracy (at object level and category level) when classification confidence over more adjacent frames is fused. On the horizontal axis the number of frames fused (temporal window). When end-of-sequence reset is not available using long temporal windows can lead to dangerous drifts (see the orange curve). }
  \label{img:seq-results}
\end{figure}

To improve classification accuracy, instead of classifying single frames, a set of temporally adjacent frames can be fused. To this purpose we implemented a simple sum-rule fusion at confidence level. The graph in Figure \ref{img:seq-results} shows the result for the Mid-VGG model. For each classification experiment (object level and category level) we tested two cases: \textit{i}) we concatenate frames from all test sequences without considering end-of-sequence events (reset); \textit{ii}) we assume that a reset signal is available. In the former, as the window size increases the risk of fusing frames from different classes increases as well. In general, fusing 40-50 frames (about 2 seconds of video) seems to be a good compromise even when sequences cannot be reliably segmented. 

\section{Continuous Object Recognition Benchmark}
\label{sec:cont-bench}

For all the continuous learning scenarios (NI, NC, NIC) we use the same test set composed of sessions \#3, \#7 and \#10. The remaining 8 sessions are split in batches and provided sequentially during training. Since the batch order can affect the final result, we compute the average over 10 runs where the batches are randomly shuffled. Moreover, for each scenario, we provide the accuracy of the cumulative strategy (i.e., the current batch and the entire previous ones are used for training) as a target \footnote{To reduce computations, the number of runs for the NC and NIC cumulative strategy is reduced to 5 and 3 respectively.}. We do not use the term upper bound because in principle a smart sequential training approach could outperform a baseline cumulative training. 
In the following sections we report results only for object level classification task (the most difficult one), since both experiments lead to the analogous conclusions. Furthermore, in this study the models were trained on RGB data only (no depth information).

\subsection{NI: New Instances}
\begin{figure}[h]
  \centering
  \includegraphics[width=0.85\textwidth]{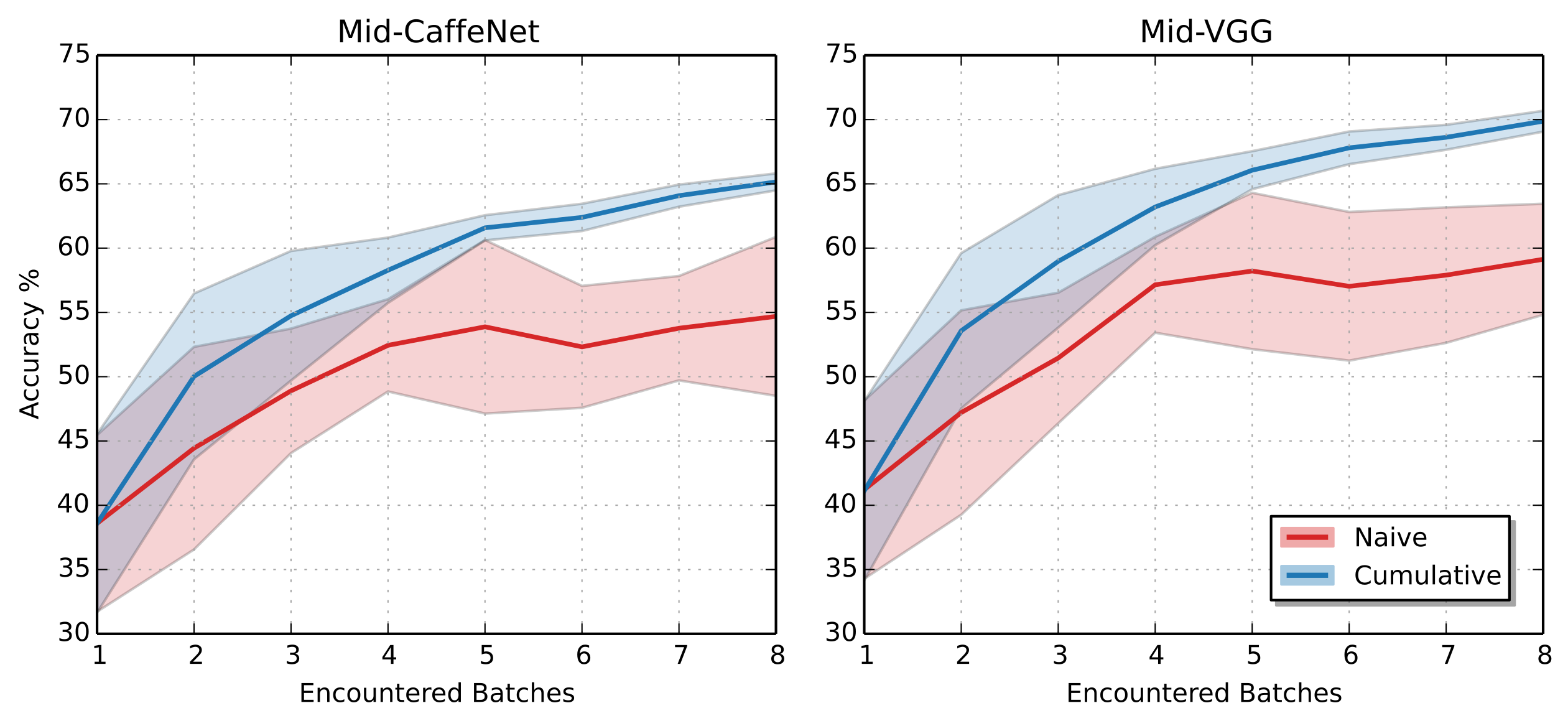}
  \caption{Accuracy results (averaged on 10 runs) for the naïve and cumulative strategies for Mid-CaffeNet and Mid-VGG. Colored areas represent the standard deviation of each curve. Tabular data available at \url{http://vlomonaco.github.io/core50}.}
  \label{img:ni}
\end{figure}

In this scenario the training batches coincides with the 8 sessions available in the training set. In fact, since each session includes a sequence (about 300 frames) for each of the 50 objects, training a model on the first session ad tuning it 7 times (on the remaining 7 sessions) is in line with NI scenario: all the classes are known since from the first batch and successive batches provide new instances of these classes to refine and consolidate knowledge.
A baseline naïve approach for this scenario is simply continuing the SGD training as new batches become available. In Figure \ref{img:ni} we compare the baseline and cumulative approaches.

The accuracy gap between the naïve and the cumulative approach is here quite modest. In fact, with a careful tuning of the learning rate and number of iterations (early stopping), forgetting can be tamed in this scenario where the model memory is regularly refreshed with new poses (scale, view angle, occlusion, lighting, etc.). Similar findings have been reported for \textsl{NORB}, \textsl{COIL100} and \textsl{BigBrother} in the NI scenario \cite{Maltoni2016-icpr} \cite{Lomonaco2016}.

\subsection{NC: New Classes}

In this scenario, for each sequential batch, new objects (i.e. classes) to recognize are presented. Each batch contains the whole training sequences (8) of a small group of classes, and therefore no memory refresh is possible across batches. In the first batch we include 10 classes, while the remaining 8 batches contain 5 classes each. To slightly simplify the task, in each of the ten runs we randomly chose the classes with a biased policy which privileges maximal categorical representation (i.e., spreading of objects of the same category in different batches). 
Defining an optimal testing protocol for NC scenario is not obvious. We initially considered three alternatives:
\begin{enumerate}
\item \textbf{NC – Partial Test Set}: at each evaluation step (i.e., after each training batch) the test set includes only patterns of the classes already presented to the network. 
\item \textbf{NC – Full Test Set}: the test set is fixed and includes patterns of all the classes. Except for the last evaluation step, the model is (also) required to classify patterns of never seen classes.
\item \textbf{NC – Full Test Set with Rejection Option}: the test set is fixed and includes patterns of all classes, however the model has the possibility to reject a pattern if it believes the pattern does not belong to any of the known classes. Since the training set does not include “negative” examples we cannot add an extra neuron for the “unknown” class, and the rejection mechanism has to compare the max class probability with a given threshold.  
\end{enumerate}

Option 1. has the drawback that as we increase the number of classes in the test set the task becomes more complex and it is difficult to appreciate the learning trend and benefits of subsequent batches. Option 3. is the most realistic one for real applications but evaluation and comparison of different techniques is more difficult because at each step instead of a single point we have a ROC curve (accuracy also depends on the threshold). Considering that our aim is comparative evaluation among continual object recognition approaches we believe that option 2. is a good trade-off between simplicity and usefulness for the task. This option also maintains the test set coherent across all scenarios (NI, NC and NIC) so we decided to adopt it.

\begin{figure}[h]
  \centering
  \includegraphics[width=0.85\textwidth]{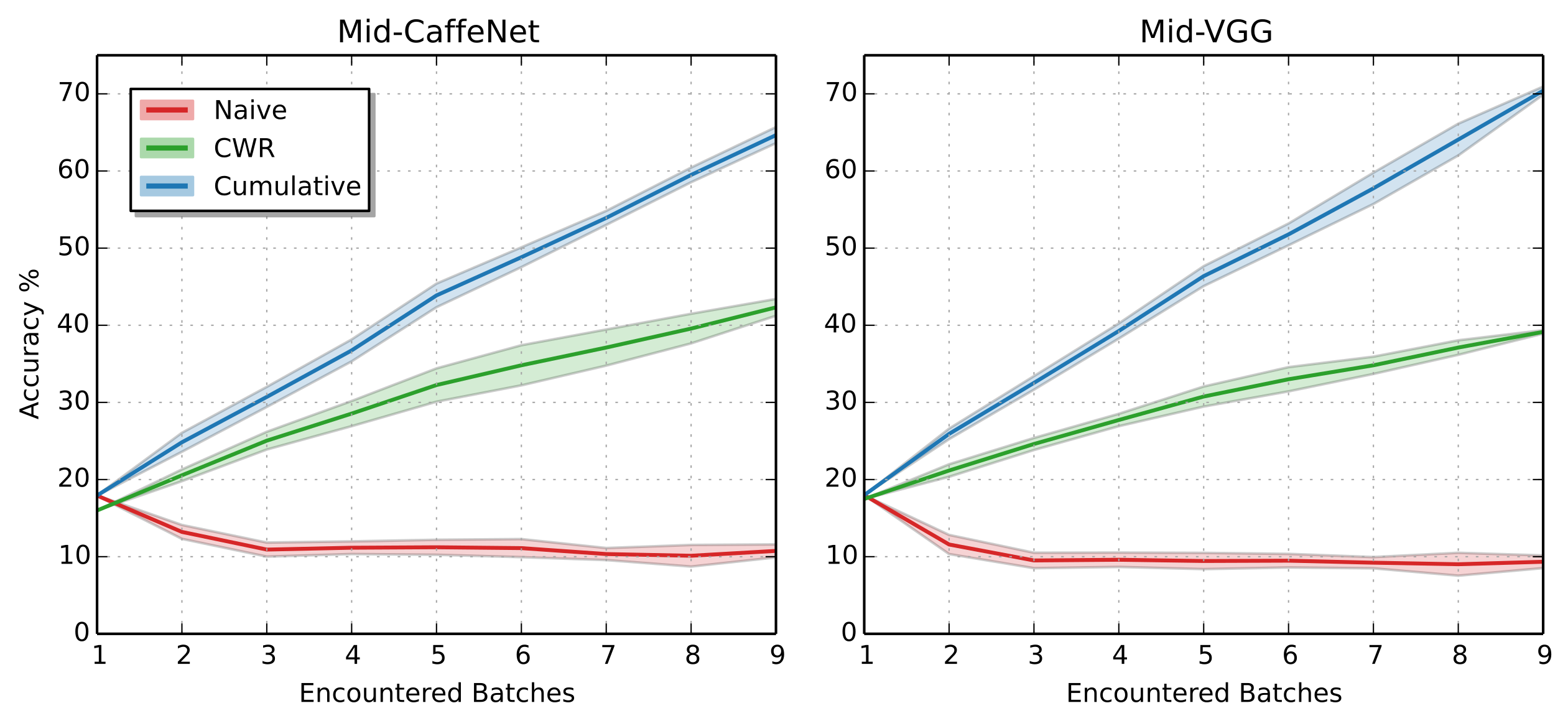}
  \caption{Mid-Caffe and Mid-VGG accuracy on NC scenario (average over 10 runs). Cumulative and Naïve approach are full depth models, while \textsl{CWR} lacks \textit{fc6} and \textit{fc7}. Colored areas represent the standard deviation of each curve. Tabular data available at \url{http://vlomonaco.github.io/core50}.}
  \label{img:nc}
\end{figure}

The naive approach in this scenario is not working at all. Graphs in Figure \ref{img:nc} show that the models completely forget the old tasks while learning the new classes: the initial accuracy drop is due to the larger size of the first batch w.r.t. the following ones. So we investigated other approaches and find out one (denoted as \textsl{CWR}: \textit{CopyWeights with Re-init}) that, in spite of its simplicity, performs fairly well and can be used as baseline for further studies (see Figure \ref{img:nc}). 

In \textsl{CWR} we skip layers \textit{fc6} and \textit{fc7} and directly connect \textit{pool5} to a final layer \textit{fc8} (followed by \textit{softmax}) while maintaining the weights up to \textit{pool5} fixed. This allows isolating the subsets of weights that each class uses. During the training two sets of weights are maintained by the model for the \textsl{pool5} $\rightarrow$ \textit{fc8} connections: \textbf{cw} are the consolidated weights used for inference and \textbf{tw} the temporary weights used for training; \textbf{cw} are initialized to 0 before the first batch, while \textbf{tw} are randomly re-initialized (default Gaussian distribution with std = 0.01, mean = 0) before each training batch. At the end of each batch training, the weights in \textbf{tw} corresponding to the classes in the current batch are copied in \textbf{cw}: this is trivial in this scenario because of the class segregation in different batches. In other words, \textbf{cw} can be seen as a sort of hippocampus where consolidated concepts are maintained, while \textbf{tw} as a short term working memory in the cortex used to learn new concepts without interfering with stable ones (see the interesting discussion in \cite{Li2016}).

\begin{figure}[h]
  \centering
  \includegraphics[width=0.55\textwidth]{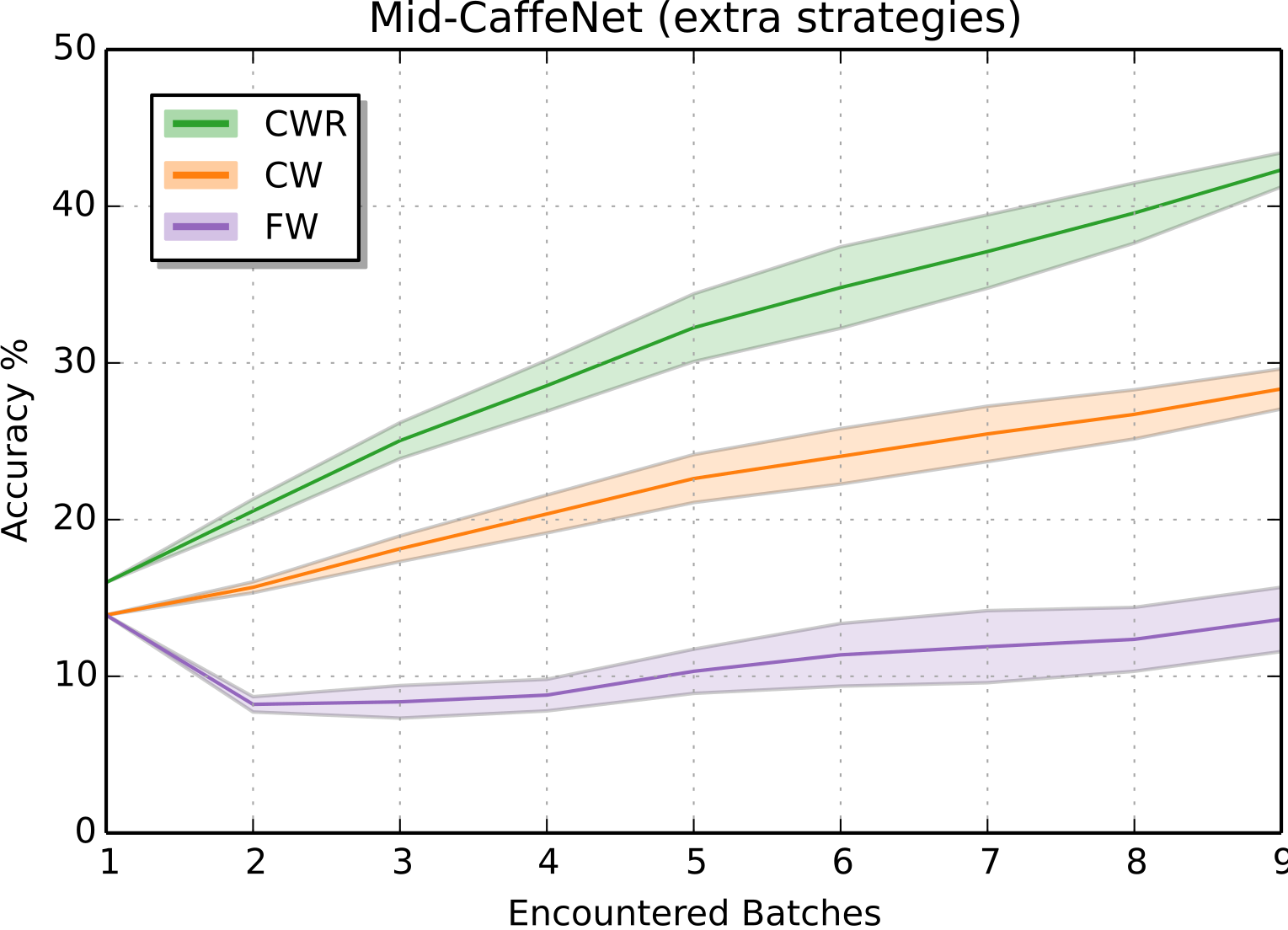}
  \caption{CWR compared with some variants: \textsl{FW} and \textsl{CW}. Colored areas represent the standard deviation of each curve.}
  \label{img:nc_extra}
\end{figure}

Other simple approaches have been implemented and tested such as: \textit{i}) \textsl{FW} where \textbf{cw} is not used and we just freeze in \textbf{tw} the class weights of the already encountered classes; \textit{ii}) \textsl{CW} which is the same of \textsl{CWR} but without the re-init step; however as shown in Figure \ref{img:nc_extra} the results obtained are significantly worse than \textsl{CWR}.

\subsection{NIC: New Instances and Classes}

In the third and last scenario, both new classes and instances are presented in each training batch. This scenario is the closest to many real-world applications where an agent continuously learns new objects and refines the knowledge about previously discovered ones.
As for NC scenario the first batch includes 10 classes, and the subsequent batches 5 classes each. However, only one training sequence per class is here included in a batch, thus resulting in a double partitioning scheme (i.e., classes and sequences). The total number of batches is 79. We maximized the categorical representation in the first batch but we left the composition and order of the 78 subsequent batches completely random. The test set is the same as in NI and NC scenarios. 

\begin{figure}[h]
  \centering
  \includegraphics[width=0.85\textwidth]{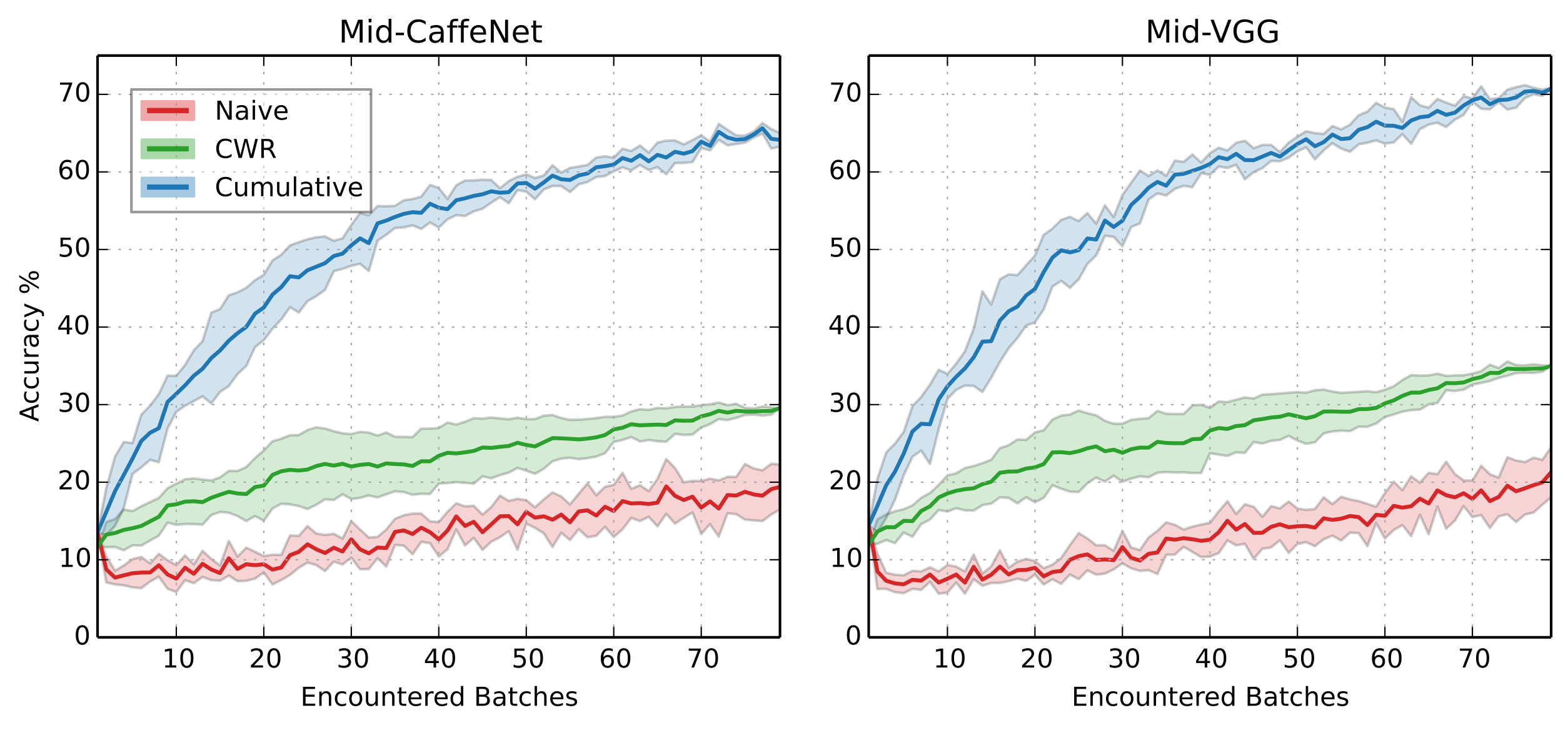}
  \caption{Mid-Caffe and Mid-VGG accuracy on NIC scenario (average over 10 runs). Colored areas represent the standard deviation of each curve. Tabular data available at \url{http://vlomonaco.github.io/core50}.}
  \label{img:nic}
\end{figure}

CWR approach for this scenario needs to be slightly adjusted. The first time a new class is encountered its \textbf{tw} weights are copied on \textbf{cw}, while at successive steps \textbf{cw} is updated as a weighted average. More precisely, for each class $i$ in the current batch:
\[
   cw[i]= 
\begin{dcases}
    tw[i]			& \text{if} \ updated[i] = 0\\
    \frac{cw[i] \cdot udpdates[i] + tw[i]}{updates[i] + 1}              & \text{otherwise}
\end{dcases}
\]
where $update[i]$ is the number of times class $i$ has been encountered so far. Figure \ref{img:nic} reports the accuracy of the naïve, cumulative and \textsl{CWR} approaches. The graph clearly shows that this scenario is very difficult (\textsl{CWR} accuracy is about half of the Cumulative approach accuracy) and there is a big room for improvements. 

\section{Conclusion}
\label{sec:conclusion}

Biological learning requires neither to store perceptual data streams nor to process them in a cumulative way; however, it effectively tackles incremental learning tasks where new object representations are continuously learned and consolidated and only the useless one are forgot. Artificial learning systems still lack these capabilities and new research is needed to fill the gap.  
In this paper we introduced a new dataset and associated benchmarks to support continuous learning studies in the context of object recognition.

As argued by many researchers our results also prove that naïve approaches such as incremental tuning cannot avoid catastrophic forgetting in complex real-world scenarios like NC and NIC. When testing new approaches on \textsl{CORe50}, the most important indicator is not the absolute accuracy on specific scenarios but the relative accuracy w.r.t. the corresponding cumulative approach. In fact, using a state-of-the-art full size CNN one can easily exceed absolute accuracy here reported for mid-size CNNs while only really effective continual learning techniques can significantly reduce the gap w.r.t. the corresponding cumulative approaches.

The proposed \textsl{CWR} approach (to be used as baseline for further studies) performs better than the naïve solution but the accuracy drops w.r.t. the cumulative approach is large and leaves much room for improvements. Directly connecting \textit{pool5} to \textit{fc8} and freezing the model parameters up to \textit{conv5}, allowed us to easily disentangle the sets of weights which are relevant for the different classes, but more sophisticate techniques are needed in presence of multiple shared layers. \textsl{EWC} \cite{Kirkpatrick2016} and \textsl{LwF} \cite{Li2016} \cite{Triki2017} are interesting approaches in this direction, and we intend to test them on \textsl{CORe50} in the near future. Other future steps in our plan are:
\begin{itemize}
\item Further extending the dataset in terms of classes and sessions;
\item Combining motion and depth to provide better object segmentation;
\item Classifying \textsl{CORe50} objects with models exploiting both RGB and depth information.
\end{itemize}

\bibliography{library}
\bibliographystyle{plain}

\appendix
\normalsize
\section{Using pre-trained CNN with different input size}
\label{appndx:mid-size}

In the recent years, the pervasiveness of deep neural networks and the complexity of training such architectures on datasets of remarkable size has led to the proliferation of pre-trained models which represent a very good starting point for many customized solutions. However, this approach requires adapting problem-specific data to a fixed size architecture which was designed and optimized to solve another task. In the context of computer vision and object recognition, for example, it is very common to stretch images of arbitrary sizes to 227 $\times$ 227 pixels which is the typical input of well-known CNN models pre-trained on \textsl{ImageNet}: this often leads to highly distort the original patterns and significantly increases inference time.
A more elegant (and efficient) approach is adapting a pre-trained model to work with input patterns of different size. This is straightforward for convolution and pooling layers thanks to local (shared) connections, but is much more problematic for fully connected layers, whose number of weights depends on the input image size. In this case, two main strategies can be used:
\begin{enumerate}
\item Applying fixed size pooling (global or pyramidal) over the last convolution/pooling layer as proposed in \cite{He2014} \cite{Ren2015} \cite{Lin2014}. However, finetuning of upper levels might be necessary if the input scale changes dramatically or the original model was not designed with a fixed-size pooling layer at all.
\item Reusing the pre-trained network up to the last convolution layer and retraining the fully connected layers from scratch on the new task and input size. A typical approach is also to train an external classifier (e.g., SVM) from pooled features just after the last convolutional layer. 
\end{enumerate}
Independently of the network adaption to a different input size, when the problem classes change, the final softmax layer needs to be replaced and re-trained from scratch.

\begin{table}[ht]
  \caption{Accuracy differences between reduced size CNNs (Mid) and the corresponding full-size models on the 50 classes task. All the models have been pre-trained on \textsl{ILSVRC-2012}. SVM training and CNN fine-tuning were performed on \textsl{CORe50}.}
  \label{tab:appendix}
  \centering
  \begin{tabular}{lp{5cm}cc}
    \toprule
     & & \multicolumn{2}{c}{\textbf{Accuracy (object level: 50 classes)}}\\
	\cmidrule{3-4}
	  & \textbf{CNN + SVM (on top of ...)} & \textbf{fc6} & \textbf{pool5}\\
	  \toprule
      1 & CaffeNet & 63,46\% & 63,14\%\\
      2 & Mid-CaffeNet &  & 52,84\%\\
      3 & VGG & 69,03\% & 70,91\%\\
      4 & Mid-VGG & & 59,25\%\\
      \midrule
      & \textbf{CNN + Finetuning} & \multicolumn{2}{c}{\textbf{Accuracy (object level: 50 classes)}}\\
      \midrule
      5 & CaffeNet & \multicolumn{2}{c}{75,97\%}\\
	  6 & Mid-CaffeNet & \multicolumn{2}{c}{65,98\%}\\
	  7 & VGG & \multicolumn{2}{c}{77,39\%}\\
	  8 & Mid-VGG & \multicolumn{2}{c}{69,08\%}\\
    \bottomrule
  \end{tabular}
\end{table}

Since in our experiments we used the classic CaffeNet and VGG models (which have not been trained in a multi-scale fashion) and we aimed at fast processing, we opted for the second strategy. Hence, we reshaped the input volume to 3 $\times$ 128 $\times$ 128, halved \footnote{As other authors we noted that such reduction has no significant impact on accuracy.} the number of units in the fully connected layers \textit{fc6} and \textit{fc7} (from 4096 to 2048) and re-trained them from scratch. This results in a relevant speedup at inference time (3.4$\times$ for CaffeNet and 4.67$\times$ for VGG). The resulting mid-size models are now suitable to be tuned on \textsl{CORe50} native 128 $\times$ 128 frames. 

Table \ref{tab:appendix} summaries our findings. For the full-size models, extracting features from \textit{fc6} or \textit{pool5} is nearly equivalent in terms of accuracy (compare columns \textit{fc6} and \textit{pool5} for raw 1 and 3 in the table). So the lack of a fully pre-trained \textit{fc6} in the mid-size models is not critical. However, in the experiments with SVM (rows 1:4), the mid-size networks loose about 10\% accuracy with respect their original version. A similar gap (just slightly smaller for VGG) can be observed when the networks are finetuned (rows 5:8).
The reason of such accuracy drop is not totally clear to us. On the one hand, if we consider finetuning experiments (rows 5:8), \textit{fc6} and \textit{fc7} have been pre-trained on a higher number of patterns in the full-size networks, and therefore it is reasonable to expect higher accuracy; on the other hand, if we consider \textit{pool5} + SVM experiments, both the network exploits the same pre-training and stretching our input patterns from 128 $\times$ 128 to 227 $\times$ 227 (in principle) does not add new information. 

We did similar experiments on other datasets (e.g., \textsl{NORB}, \textsl{COIL100}, \textsl{BigBrother}, \textsl{iCub}) and obtained close results: it seems that the zoomed image, even if a blurred, allow a more detailed feature extraction to be performed by the network. This can be due to the spatial scale of the filters learned on \textsl{ILSVRC-2012} or by a richer hierarchical representation (more neurons and link between neurons cover the object region). We believe that more investigations are necessary to fully understand the reasons and to make available pre-trained mid-size networks (for patterns whose native size is close to 128 $\times$ 128) which are competitive with full-size ones. 

\end{document}